\documentclass[journal]{IEEEtran}

\usepackage[nocompress]{cite}

\ifCLASSINFOpdf
\else
\fi

\usepackage{amsmath}

\usepackage{amssymb}
\usepackage{pgfplots}
\usepackage{booktabs}
\usepackage{comment}
\usepackage{stfloats}
\pgfplotsset{compat=1.17}

\begin{document}

\title{Knock Detection in Combustion Engine Time Series Using a Theory-Guided 1D Convolutional Neural Network Approach}

\author{\IEEEauthorblockN{
Andreas B. Ofner
, Achilles Kefalas
, Stefan Posch
, and Bernhard C. Geiger 
}

\thanks{Manuscript received April 12, 2021. \par
A. B. Ofner is with the Know-Center GmbH, Graz, Austria (e-mail: aofner@know-center.at). \par
A. Kefalas is with the Institute for Internal Combustion Engines and Thermodynamics, Graz University of Technology, Graz, Austria (e-mail: kefalas@ivt.tugraz.at). \par
S. Posch is with the Large Engines Competence Center GmbH, Graz University of Technology, Graz, Austria (e-mail: stefan.posch@lec.tugraz.at). \par
B. C. Geiger is with the Know-Center GmbH, Graz, Austria (e-mail: bgeiger@know-center.at).
}}

\markboth{IEEE/ASME Transactions on Mechatronics}%
{Shell \MakeLowercase{\textit{et al.}}: Bare Demo of IEEEtran.cls for IEEE Transactions on Magnetics Journals}

\IEEEtitleabstractindextext{%
\begin{abstract}
This paper introduces a method for the detection of knock occurrences in an internal combustion engine (ICE) using a 1D convolutional neural network trained on in-cylinder pressure data. 
The model architecture was based on considerations regarding the expected frequency characteristics of knocking combustion.
To aid the feature extraction, all cycles were reduced to 60° CA long windows, with no further processing applied to the pressure traces. 
The neural networks were trained exclusively on in-cylinder pressure traces from multiple conditions and labels provided by human experts. 
The best-performing model architecture achieves an accuracy of above 92\% on all test sets in a tenfold cross-validation when distinguishing between knocking and non-knocking cycles. 
In a multi-class problem where each cycle was labeled by the number of experts who rated it as knocking, 78\% of cycles were labeled perfectly, while 90\% of cycles were classified at most one class from ground truth. 
They thus considerably outperform the broadly applied MAPO (Maximum Amplitude of Pressure Oscillation) detection method, as well as other references reconstructed from previous works. 
Our analysis indicates that the neural network learned physically meaningful features connected to engine-characteristic resonance frequencies, thus verifying the intended theory-guided data science approach. 
Deeper performance investigation further shows remarkable generalization ability to unseen operating points. 
In addition, the model proved to classify knocking cycles in unseen engines with increased accuracy of 89\% after adapting to their features via training on a small number of exclusively non-knocking cycles. 
The algorithm takes below 1 ms (on CPU) to classify individual cycles, effectively making it suitable for real-time engine control.
\end{abstract}

}

\maketitle

\IEEEdisplaynontitleabstractindextext

\IEEEpeerreviewmaketitle

\section{Introduction}

\IEEEPARstart{E}{ngine} knock is the broadly used term for undesired stochastic phenomena in a spark ignition (SI) combustion engine's thermal process. 
Thereby, various influence factors such as increased temperature and pressure within a cylinder lead to the fuel mixture's auto-ignition before the propagating flame front induced via the spark plug \cite{MAURYA:19}. 
While the combustion itself thus can gain efficiency, the pressure waves created can lead to severe damages in the engine. 
Among those are the erosion of piston crowns, top land and cylinder head, as well as breakage of piston rings, cylinder bore scuffing and other structural damages \cite{ZHENG:12}.
Furthermore, acoustic resonances are induced, resulting in the characteristic knock sound associated with the phenomenon.
To prevent such abnormal combustion and the related risks, the ignition is retarded and the compression ratio is constrained.
Those measures result in decreased efficiency, effectively placing knock occurrence as a limiter on engine operation \cite{MAURYA:19, WANG:17}.
\newline
Hence, manufacturers have developed a variety of methods designed to avoid, minimize or even prevent auto-ignition occurrences. 
Among these are the use of high-octane, "knock-resistant" fuels \cite{LEONE:15} or more conservative spark timing calibration maps \cite{AYALA:06}. 
Considerable efforts have been invested into exhaust gas re-circulation (EGR) systems, which use a portion of the previous combustion's exhausts to serve as a coolant for the current combustion. 
The thus achieved temperature decrease effectively lowers emissions and increases efficiency while successfully creating less knock-prone conditions\cite{BOZZA:16}\cite{TEODOSIO:15}.
Recent approaches have also shown a gradual reduction of knock intensity, general combustion stability and emission of several exhaust gases when directly injecting water into the cylinder \cite{VALERO:18}. 
However, despite its positive impact, the method is connected to increased soot emissions \cite{LI:20}.
Since the mentioned methods are susceptible to consumer behavior, can cause problems with selected exhaust products or possibly prevent the engine from achieving optimal performance, respectively, a knock detection is often a preferable approach. \newline
Conventional knock detection works by comparing incoming vibration or pressure data with manually predetermined thresholds, assuming auto-ignition whenever that value is exceeded. 
Crossing these threshold values usually triggers a retardation of the ignition timing in order to re-stabilize the combustion process. 
This way, the engine can be operated at high load and performance for increased amounts of time without the need for additional physical systems, their corresponding disadvantages and the present risk of their failure.
While there exist classification methods based on more complex physics- and chemistry-inspired approaches, most of the underlying methodologies of detection mechanisms share one major drawback: They must be carefully calibrated to an engine's momentary operating conditions.
This, however, leads to most detection methods being highly customized for the engine they have been designed for - in several cases being even further adapted to one single operating point. 
A selection of these methods is further described in section \ref{sec:related}. \newline
The approach presented in this paper is designed to eliminate this dependency on concrete engine conditions, establishing a convolutional neural network (referred to as CNN) as classification method which is trained to judge from self-learned features instead of manual input parameters comparable to calibrated thresholds. 
The aim hereby is that training and dataset labeling need to be conducted only once, with the model being able to generalize with minimal to no adjustments to new engines.
To promote this, a diversified dataset - detailed in section \ref{sec:dataset} - is used for training the model. 
Additionally, its architecture is designed using the principles of theory-guided data science \cite{KARPATNE:17} to capture specific resonance frequencies obtained from physical relations, further described in sections \ref{sec:method} and \ref{sec:result}. 
\newline
Lastly, the approach proposed in this paper is designed to operate fast enough to be deployed as part of a real-time engine control system.

\section{Related Work}
\label{sec:related}

Knock detection - and engine control in general - have in the past been conducted mainly via customized methodologies to ensure efficient operation. 
Many of those predominantly rely on physics-based approaches or are calibrated to one single engine or even individual operating points of those engines. 
Recently, however, artificial intelligence solutions have started to gain track.\newline
Most commonly, traditional physics-based knock modeling relies on the approximation of phenomena occurring in the combustion process using chemical kinetics models or empirical correlations. 
A methodology proposed in \cite{TOUGRI:16} hereby uses the Arrhenius equation for describing the fuel ignition, the Vibe approach for energy release \cite{VIBE:70} and the Woschni equations for heat loss \cite{WOSCHNI:67}. 
The authors' 0D model is combined with a previous empirical auto-ignition approach \cite{YATES:08} and calibrated with simulation data. 
The model is working successfully for the specific operating condition it was designed for.\newline
Netzer et al. \cite{NETZER:17} on the other hand propose a chain of chemical and physical models which next to detecting auto-ignition also estimates its severity. 
The model is predominantly built on fuel characteristics and calibrated to deployment at the knock limit.\newline
In the work of Bevilacqua et al. \cite{BEVILACQUA:17}, a 3D Computational Fluid Dynamics approach is employed. 
The resulting model uses the expanded focus of the three-dimensional method to also incorporate geometric effects within the engine to evaluate knock risk. 
For the purpose of validation, the approach is calibrated to fit two operating conditions: low-end torque and peak power operation.\newline
Treating the broader problem of engine control for HCCI engines using artificial intelligence, \cite{VAUGHAN:15} presents a weighted-ring extreme learning machine for predicting heat release related quantities. 
The model is pre-trained on offline data and adapting to current engine conditions via online re-training and is proven functional in real applications. 
As input, the algorithm processes a 6-dimensional vector, capturing several pressure values within a cycle, the start of ignition, the injection pulse width and current heat release data.\newline
Other work, however, argues against the use of heat release data for knock detection \cite{SHAHLARI:12}. 
They further recommend against TVE (Threshold Value Exceeded) methods due to their delay in recognizing a knock occurrence. 
The authors present an expression and supporting experimental evidence for computing the oscillation frequencies likely to occur in knocking combustion from the engine geometry. 
The proposed SEPO (Signal Energy of Pressure Oscillation) method, however, is described as suitable for a posteriori diagnostic evaluations only. \newline
A customized approach based on a similar SER (Signal Energy Ratio) method is able to correct the aforementioned detection delay in TVE methods \cite{KIM:15}. 
Using a median and smoothing filter to prevent eventual biases from Butterworth-type filters, the author determines a new threshold calculated as the sum of mean and the quintupled standard deviation from 5 °CA before the knock onset found by regular TVE methods.\newline
Both aforementioned methods, however, can show significant errors or missing categorization as indicated in \cite{CHO:19}. 
Their own approach uses a 5-layer fully connected neural network, trained on approximately 30,000 cycles from one engine - 12,500 of which having previously been classified as knocking via a fixed threshold MAPO (Maximum Amplitude of Pressure Oscillation) criterion.
The model's performance is measured against a custom reference method, which is a manual placement of labels based on observed sudden increases in pressure traces. 
Compared to \cite{SHAHLARI:12}, \cite{KIM:15} and a TVE method, the authors' deep learning model shows improved RMSE (Root Mean Square Error) score and faster processing speed than the former two approaches. \newline
In the work of Panzani et al. \cite{PANZANI:19}, two methodologies for detecting knock occurrences are introduced - both based on a PCA (Principal Component Analysis) of the in-cylinder pressure signal. 
While a previous study from the same core-group of authors \cite{PANZANI:16} uses only the first three thus extracted principal components, their more recent paper suggests calculating up to 20 principal components per cycle as a fundamental step. 
Subsequently, the inner products between these principal components and the pressure trace under consideration are used as input parameters for a logistic regression algorithm to classify knock. 
This procedure, labeled as "data-driven" method achieved a stable maximum accuracy of approximately 87\% during classification when using 5 or more principal components. 
As an alternative, the authors introduce the "Eigenpressure" methodology. 
Here, the principal components are used to reconstruct the original pressure signal. 
Due to the lost information from the dimensionality reduction process, the resulting curves represented a smoother version of the input signal.
By subtracting the newly acquired signal from the original in-cylinder pressure trace, the authors can isolate highly dynamic oscillations from the input signal, thereby effectively replacing a band-pass filter.
An applied MAPO criterion on this "residual" signal is then used as one of two features in a  logistic regression algorithm. 
The second feature is the RMSE of the reproduced pressure trace compared to the original.
In the paper, this "Eigenpressure" method achieves a maximum accuracy of approximately 93\% using 10 principal components, which is equivalent to the performance of a standard MAPO procedure applied on the authors' data.\newline
Another recently published approach \cite{SHIN:20} also employs a 7-layer deep, fully-connected neural network and achieves outstanding accuracy. 
Similar to previously referenced methods, they use a multi-dimensional feature vector as input for their model. 
It contains engine speed, spark timing and throttle angle to describe the current operating condition, and intake temperature and pressure as sensor-acquired parameters. 
However, all data used for training and testing purposes was produced via 1D simulation tools. 
Furthermore, the model's evaluation is conducted for 4 distinct datasets, the biggest of which comprising a total of 480 cycles before train-test-split. 
A tenfold cross-validation is described, however, there is no apparent mix between data points from different sets and no information on the distribution of knocking and non-knocking cycles during the cross-validation process. \newline
Siano et al. use an approach based on the DWT (Discrete Wavelet Transform) of engine block vibration data \cite{SIANO:15}. 
Their dataset comprises 1,200 total cycles from 3 distinct operating points. 
While proving that their method is more sensitive than the established MAPO criterion, it still relies on a threshold. 
Furthermore, the authors describe improvements in detection speed as necessary before deployment in real applications. Recent work is combining the wavelet transform with convolutional neural networks to achieve remarkable accuracy levels \cite{kefalas2021detection}. \newline
As another method based on frequency analysis, Bares et al. describe a comparison of two FFT integrals as their knock detection approach \cite{BARES:18}. 
These FFTs process 24 °CA wide Blackman-Harris windows applied to a filtered pressure signal, centered at the points of maximum heat release rate and maximum temperature of unburned gas, respectively. 
A knock is detected whenever the latter integral outweighs the former.
With this approach, the authors aim to correct a common shortcoming of the MAPO method, where resonances caused by regular combustion exceed the threshold and the corresponding cycles are classified as knocking.\newline
In a general attempt using machine learning techniques on time series data, \cite{ZHANG:18} shows successful implementation of a feedforward, fully-connected neural network for anomaly detection. 
Their method also proves the possibility to forego a separate feature extraction step and instead taking the actual time series as input.
Despite their work being based on key performance indicators of internet companies, insights have been valuable for the initial network design conducted for this paper. \newline
Building on that approach, \cite{MENG:20} describes an Auto Encoder for anomaly detection using convolutional layers to extract learnable characteristics from time series data. 
As a guideline for the present work, elements from both of the last-named methodologies were unified for a new, customized approach designed for maximum efficiency for the nature of the underlying problem.\newline
The success of physics-based and data-driven methods, respectively, for only a narrow scope of the underlying problem is discussed by Karpatne et al. in their paper on theory-guided data science \cite{KARPATNE:17}. 
While arguing that both approaches represent extremes on the spectrum of possible solution strategies, the authors provide concepts of synergies between domain knowledge and data science. 
In regard of the proposed methodologies, the models presented in this paper follow a hypothesis merging insights from physics-based theories with the design of CNN architectures.\newline
The presented work aims at leveraging the potentials observed in time series classification with neural networks to establish a new knock detection approach, which is (i) fast enough to be operated in real-time engine control, (ii) able to generalize to new operating points without extensive reparameterization and (iii) able to achieve precise rating from raw pressure data, without any form of pre-processing.  

\section{Dataset Description}
\label{sec:dataset}

For the work described in this paper, solely in-cylinder pressure was measured as input for the CNN model. 
Labels were provided by a committee of experienced engineers.
It was decided that working with in-cylinder pressure sensors was the preferable option when compared to engine block vibration sensors, as the latter showed a strong dependence on sensor positioning and high levels of noise. Furthermore, previous investigations conclude that in-cylinder pressure traces lead to the most precise detection results \cite{BARES:18}.\newline 
Data was collected from a variety of large, single-cylinder test bench engines, the broad specifications of which are given in Table \ref{tab:engdata}. 
Due to collaboration with an industrial partner, the precise values have to be concealed for reasons of confidentiality, with the exception of the bore size which serves the approach's underlying hypothesis.
Hence, the table primarily serves to highlight distinctions and similarities between the engines. 
The engines in this study are designed to work as generators for the specifications of the European electricity grid and were run at an engine speed of 1,500 revolutions per minute, which is the generators' standard operation rate for the European grid. 
In order to gain a certain independence from the engine's rotational speed, all measurement data was translated to the crank angle (CA) domain, with the Top-Dead-Center position (TDC) as central reference point.

\begin{table}[ht!]
    \centering
        \caption{Specifications of analyzed engines}
         \resizebox{0.48\textwidth}{!}{%
        \begin{tabular}{lcccc}
        \toprule
        & \textbf{Units} & \textbf{Engine A} & \textbf{Engine B} & \textbf{Engine C}\\
        \midrule
        Bore & [mm] & 145 & 145 & 190 \\
        Cylinder Head & [-] & Head A & Head B & Head C \\
        Piston & [-] & Piston A & Piston A & Piston C \\
        Compression Ratio & [-] & Ratio A & Ratio B & Ratio C \\
        Spark Plug & [-] & Plug A & Plug B & Plug C \\
                Pre-Chamber & [-] & None & Chamber A & Chamber B\\
        \midrule
        No. of Operating Points & [-] & 14 & 15 & 9 \\
        Cycles per OP & [-] & 60 & 100 & 60 \\
        BMEP range & [bar] & 1-8 & 5-18 & 9-13\\
        Ignition Timing & [°bTDC] & 16-24 & 18-22 & 20 \\
        Lambda & [-] & 1.0-1.6 & 1.1-1.5 & 1.3-1.9\\
        \midrule
        \textbf{Total cycles} & [-] & \textbf{840} & \textbf{1500} & \textbf{540} \\
        Knock/No-knock ratio & [-] & 306/534 & 1077/423 & 202/338 \\
        \bottomrule
        \end{tabular}%
        }
    \label{tab:engdata}
\end{table}

Judging from the characteristic parameters, it can be concluded that engines A and B are generally similar to each other, with the distinct difference lying in the presence of a pre-chamber. 
Engine C on the other hand can be distinguished by various factors, all of which are mainly due to its increased size.\newline
All of the operating points making up the subsets comprise raw and unfiltered in-cylinder pressure data over CA starting at 360.0 °CA before TDC position in the ignition stroke until 359.9 °CA after, covering all four strokes of a regular combustion engine thermal cycle. 
The resolution is 0.1 °CA for each cycle, effectively resulting in time series blocks of 7200 data points for each cycle and a sampling frequency of 90 kHz for the given rotational speed. 
This is considered sufficiently high for the target frequencies investigated in this paper (see Section \ref{sec:method}). 
 \newline
Subsets A, B and C formed the basis for train and test data in building and optimizing the CNN model, adding up to a total of 2,880 cycles recorded over 38 distinct operating points in three different engines. 
This data composition was chosen in order to form a more diverse learning base and thus amplify the model's ability to generalize.

\subsection{Cycle labels}

A major hindrance for the application of supervised machine learning methods in knock detection is the uncertainty of labels acquired via traditional sources such as designated knock sensors. 
The labels used for the CNN's supervised learning process were provided by five experts in cycle analysis affiliated with the Large Engines Competence Center GmbH Graz. 
All 2,880 cycles were classified separately by each expert, using their experience as well as any number of post-processing tools of their choice. 
Results from other experts were not available to any of the judges during the labeling process. 
Among the criteria analyzed by the experts for cycle rating were: FFT, harmonics, heat release, cycle-to-cycle variations, amplitude jumps and ratios or high-frequency pressure signal evaluation.
As a result of this procedure, a (2,880$\times$5) matrix was obtained containing a binary rating of either 0 (normal combustion) or 1 (knocking combustion) for each of the cycles in the full training data set. \newline

\begin{figure}[!ht]
\centering
\resizebox{0.3\textwidth}{!}{%
    \begin{tikzpicture}
        \begin{axis}[ybar interval, mark=no, ymin=0, ymax=1200,  xmin=-0.5, xmax = 5.5, enlargelimits=false, xlabel={Relative label}, ylabel={Number of cycles}, xtick pos=bottom, ytick pos=left, grid=none, xtick={-0.5,0.5,1.5,...,5.5}, xticklabels={0,1,2,...,5}, extra x ticks = 2.5, extra x tick labels={}, extra x tick style={grid=major,major grid style={thick,draw=black}}]
        \addplot [fill=lightgray] plot coordinates { (-0.50, 803) (0.5, 250) (1.5, 232) (2.5, 137) (3.5, 325) (4.5, 1123) (5.5, 0) };
        \end{axis}
    \end{tikzpicture}%
    }
    \caption{Distribution of relative labels attributed to individual cycles by experienced test bed operators and engineers. The solid vertical line indicates the decision boundary for binary labels "normal" (left from line) and "knocking" (right from line)}
    \label{fig:dist_label}
\end{figure}
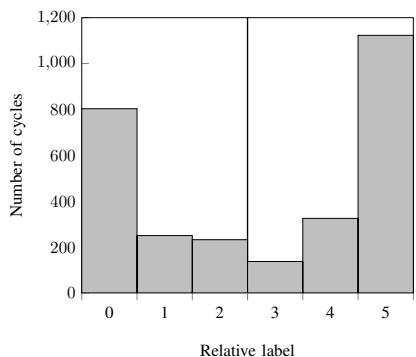
Two sets of labels were obtained from this process, hereafter referred to as the 'relative' labels and the 'binary' labels, respectively.
The set of relative labels was obtained by summing the experts' ratings for each cycle. 
Consequently, depending on their respective label, cycles were interpreted as "severe knock" (5), "knock" (4) and "light knock" (3), while those with lower values (2 \& 1) were judged as non-damaging cycles, and "normal combustion" (0). \newline
To arrive at the binary label, a simple majority rule was applied. 
Summing the ratings for each line, any value of 2 and below indicated a "normal" rating while 3 and above lead to a classification of "knocking". 
Fig. \ref{fig:dist_label} shows the distribution of votes for the 2,880 cycle data set. 
\noindent As can be seen in Fig. \ref{fig:dist_label}, based on their subjective methods, the experts achieved consensual ratings ("normal combustion" or "severe knock") for 1,926 of 2,880 cycles ($\approx 66,9\%$).

\section{Methodology}
\label{sec:method}

\subsection{Pre-processing}
The aim of pre-processing was to reduce the full cycle to a window of maximum relevance for the knock phenomenon. 
After several investigations, it was decided to cut out a 60° CA window from TDC position to 60° CA aTDC.
Hence, the peak pressure as well as the vital parts of the combustion stroke were isolated for processing with the neural network approach.
As a side effect, this also increases the classification speed of the model, as fewer parameters have to be propagated through the network.
Apart from this window slicing, no other processing such as, e.g. scalers were applied to the data.

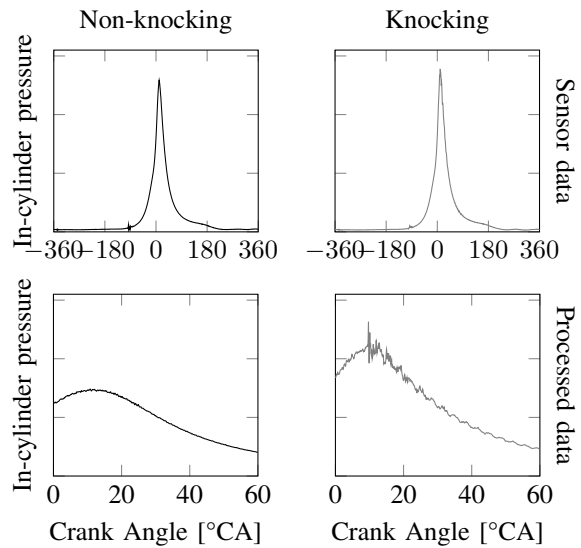
\begin{figure}[!ht]
    \centering
    \begin{tikzpicture}
        \begin{axis}[width=4.3cm, height=4cm, ymin=0, ymax=155,
        xmin=-360, xmax=360, xlabel={Non-knocking}, ylabel={In-cylinder pressure}, xtick={-360,-180,...,540}, yticklabels={,,},
        every axis y label/.style={at={(current axis.west)}, rotate=90, above=0.1cm},
        every axis x label/.style={at={(current axis.north)}, above=0.1cm}]
        \addplot[color=black,smooth] table [x=a, y=b, col         sep=semicolon]{fig/FIG2_src_A.csv};
        \end{axis}
    \end{tikzpicture}
    \begin{tikzpicture}
        \begin{axis}[width=4.3cm, height=4cm, at={(0,0)}, ymin=0, ymax=155, xmin=-360, xmax=360, xlabel={Knocking}, ylabel={Sensor data}, xtick={-360,-180,...,540}, yticklabels={,,}, every axis y label/.style={at={(current axis.east)}, rotate=-90, above=0.1cm}, every axis x label/.style={at={(current axis.north)}, above=0.1cm}
            ]
        \addplot[color=gray,smooth] table [x=a, y=d, col     sep=semicolon]{fig/FIG2_src_A.csv};
        \end{axis}
    \end{tikzpicture}
    
    \begin{tikzpicture}    
        \begin{axis}[width=4.3cm, height=4cm, ymin=0, ymax=155, xmin=0, xmax=600, xlabel={Crank Angle [°CA]}, ylabel={In-cylinder pressure},  xtick={0,200,...,700}, xticklabels={0,20,...,70}, yticklabels={,,},
        every axis y label/.style={at={(current axis.west)}, rotate=90, above=0.1cm}]
        \addplot[color=black,smooth] table [x=a, y=b, col         sep=semicolon]{fig/FIG2_src_B.txt};
        \end{axis}
    \end{tikzpicture}
    \hspace{0.25cm}
    \begin{tikzpicture}
        \begin{axis}[width=4.3cm, height=4cm, at={(0,0)}, ymin=0, ymax=155, xmin=0, xmax=600, xlabel={Crank Angle [°CA]}, ylabel={Processed data}, xtick={0,200,...,700}, xticklabels={0,20,...,70}, yticklabels={,,}, every axis y label/.style={at={(current axis.east)}, rotate=-90, above=0.1cm}
             ]
        \addplot[color=gray,smooth] table [x=a, y=b, col     sep=semicolon]{fig/FIG2_src_C.txt};
        \end{axis}
    \end{tikzpicture}
    \caption{Non-knocking and knocking cycles before (top) and after pre-processing (bottom)}
    \label{fig:preprocessed}
\end{figure}
The model therefore uses the raw in-cylinder pressure signal. 
Fig. \ref{fig:preprocessed} shows examples for full knocking and non-knocking cycles (top) and the pre-processed windows (bottom).

\subsection{Customized train-test-split}

\begin{figure}[!ht]
    \centering
    \includegraphics[width=8.8cm]{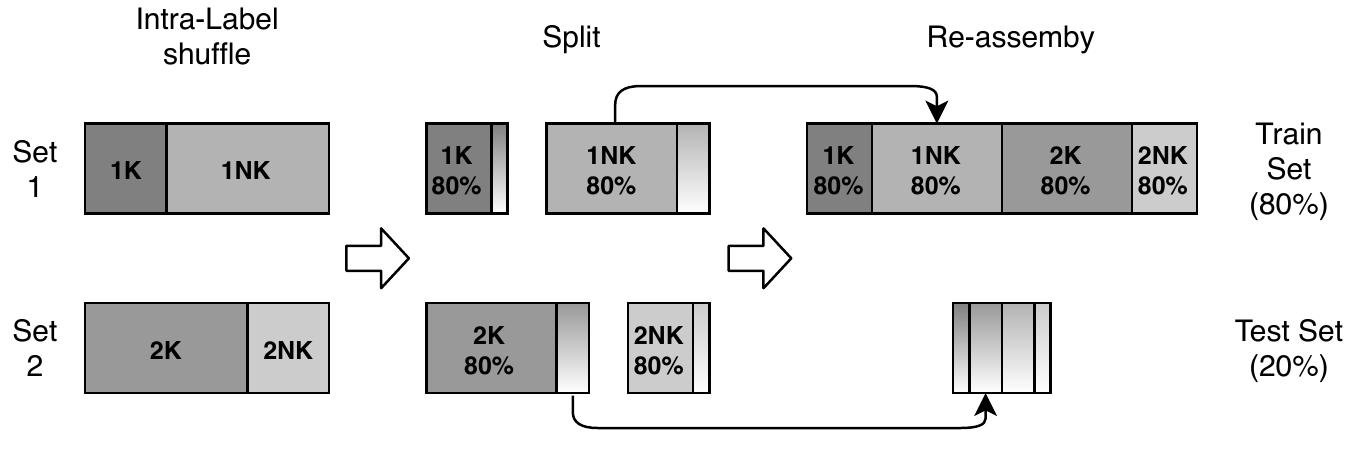}
    \caption{Train-Test-Split method. K indicates knocking, NK indicates non-knocking.}
    \label{fig:setsplit}
\end{figure}

Another factor to consider in pre-processing is the train-test-split. 
Due to the uneven distribution of knocking and non-knocking cycles within subsets A, B and C and the fact that knocking appears to come in multi-cycle bursts, i.e., same label sequences, a regular split would tend to have a majority of either knocking or non-knocking cycles. 
To ensure distribution similar to the one in the complete subset, a custom split function was written.\newline
This method first creates one subset for each label from the complete 2,880 cycle dataset before shuffling the cycles. 
The shuffling is implemented to prevent certain operating points from being entirely omitted from the training set. 
Following this process, the sets are each split in a ratio given by the user, e.g. a 70/30 split. 
Then, corresponding subset splits are put back together to create the training and test set, respectively. This is illustrated in Fig. \ref{fig:setsplit} for a 80/20 split.

\noindent Since train and test sets in this work consist of individual splits from three different subsets, the notation of splits will describe the overall split as numeric description of the subsets' share in the training sets. 
A split where, e.g. $70\%$ of subset A, $50\%$ of subset B and $45\%$ of subset C are included in the training set, would be denoted as a $70/50/45$ split. 
Unless noted otherwise, all of the results shown in later sections have been achieved using a $70/70/70$ split (70\% of each subset in the training set).\newline

\subsection{Model architecture}

As described in section \ref{sec:related}, previous work \cite{SHAHLARI:12, BARES:18} has shown that knock occurrences can be connected to certain resonance frequencies observed in the pressure traces. 
More specifically, the authors computed these resonance frequencies as:

\begin{equation}
    f = a \cdot \frac{B}{\pi D_{b}}
    \label{eq:01}
\end{equation}

\noindent with $a$ representing the speed of sound, estimated at 966 $\frac{m}{s}$ for a maximum combustion chamber temperature of 2500 K, $B$ being the vibration mode factor as extracted from Bessel's equations, $D_b$ being the bore diameter in millimeters and $f$ giving the mode's frequency in kHz \cite{DRAPER:38} - yielding the modes visible in Table \ref{tab:modes}.

\begin{table}[!ht]
    \centering
    \caption{Acoustic mode frequencies per engine bore in kHz}
    \resizebox{0.45\textwidth}{!}{%
    \begin{tabular}{cccccc}
    \toprule
      & \textbf{1st circ.} & \textbf{2nd circ.} &  
     \textbf{1st rad.} & \textbf{3rd circ.} & \textbf{1st comb.} \\ 
     \textbf{$D_b$ [mm]} & a = 1.841 & a = 3.054 & a = 3.831 & a = 4.201 & a = 5.318 \\
     \midrule
    65 \cite{SHAHLARI:12} & 8.7 & 14.4 & 18.1 & 19.9 & 25.2 \\
    145 & 3.9 & 6.5 & 8.1 & 8.5 & 11.3 \\
    190 & 3.0 & 4.9 & 6.2 & 6.5 & 8.6 \\
    \bottomrule
    \end{tabular}%
    }
    \label{tab:modes}
\end{table}

Applying the formula \eqref{eq:01} shows considerably lower frequencies for the acoustic modes in large engines. 
However, higher order circular and combined modes fall into similar ranges as smaller engines' lower order modes. 
In order to capitalize on the presumed frequency spectrum peaks, the CNNs' architectures were build on the hypothesis that a network's first layer can be directly tuned to detect them more efficiently by adapting the filter size parameter. 
Hereby, the filter or kernel size $k$ was adapted to be equal the period $T$ of one oscillation at a specific target frequency $f_{t}$ in the above mentioned range of interest.\newline
To initiate the process, the in-cylinder pressure data's underlying rotational speed of $1500$ RPM has to be converted to the crank angle domain, equaling 9000 °CA per second.  
Hence, considering a target frequency of 3000 Hz - the calculated 1st circular mode of a 190mm bore engine (see Table \ref{tab:modes}) - and its corresponding time period of $3.3 \cdot 10^{-4}$ seconds - leads to the crank angle window for one recorded oscillation:

\begin{equation}
    9000\text{ °CA/s} \cdot \frac{1}{3000s^{-1}} = 3\text{ °CA}
    \label{eq:window}
\end{equation}

\noindent Given the signal resolution of $0.1$ °CA, this results in a final kernel size of $k = 30$ for a filter designed to target vibrations at 3000 Hz.\newline
Since the kernel size parameter is limited to natural numbers, one instance can be the best approximation for a certain frequency range.
This range will broaden with increasing frequency, as the results of equation \eqref{eq:window} move closer together as the time period decreases.
Table \ref{tab:kernelsize} gives an overview of the parameters implemented in each model. \newline
Models $a$, $b$ and $c$ were aimed towards large engine characteristic frequencies, with the specific target frequency values in agreement with the calculated acoustic mode frequencies in Table \ref{tab:modes}. 
Model $d$ was tuned to fit small engine lower order modes. 
Due to the effect described above, the hence chosen kernel size also represents a good approximation for target frequencies of higher order modes in large engines.

\begin{table}[!ht]
    \centering
    \caption{Model parameters for different target frequencies}
    \resizebox{0.3\textwidth}{!}{%
    \begin{tabular}{ccc}
    \toprule
         & \textbf{Kernel size} & \textbf{$f_t$ range [kHz]}  \\ \midrule
         Model a & 30 & 3.0 \\
        Model b & 23 & 3.8 - 4.0 \\
        Model c & 18 & 4.8 - 5.2 \\
        Model d & 11 & 7.8 - 8.7 \\
        \bottomrule
    \end{tabular}%
    }
    \label{tab:kernelsize}
\end{table}

Including this intitial, customized layer, each of the models described in this paper were built with three convolutional layers, followed by two fully-connected layers.
This rather "shallow" architecture was chosen to effectively reduce the amount of learnable parameters in the neural network which is crucial for enhancing processing speed and the model's ability to efficiently learn features even from smaller training sets.
Bias terms were omitted for all convolutional layers, but were activated for both fully-connected layers.\newline
In order to increase the number of extracted features from each instance within the model, each convolutional layer can employ an arbitrary number of same-sized kernels - henceforth referred to as "channels" \cite{ALBAWI:17}.

\begin{figure*}[!ht]
    \centering
    \includegraphics[width=0.9\textwidth]{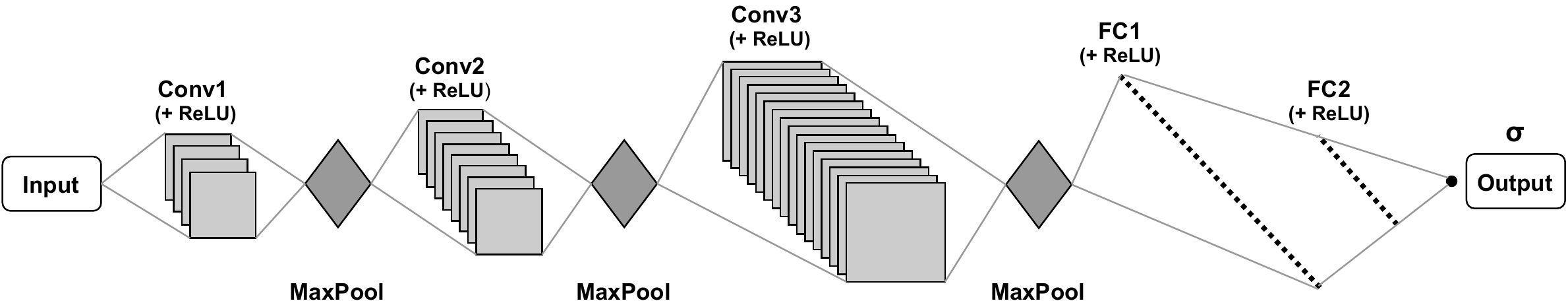}
    \caption{Fundamental architecture for all CNN models. The first two convolutional layers share a kernel size, while the second layer doubles the number of channels. The last convolutional layer doubles both kernel size and number of channels. Pooling layers follow each convolutional layer. The first fully connected layer reduces the remaining flattened feature vector to half its size, while the final fully-connected layer reduces it to one decisive parameter which is scaled via a Sigmoid function. All other layers use a ReLU activation function.}
    \label{fig:cnn_arch}
\end{figure*}

\noindent The implementation for this study follows the common design of small kernels and fewer channels at the start of the network, with kernel size and number of channels progressively increasing \cite{ALBAWI:17}.
The stride parameter for each convolutional layer was set to the minimum value of 1, while a padding of 5 was applied for each convolutional layer in order to properly process the pressure window's start and end.
This definition of the first three layers leads to a large increase in parameters before the feature vectors proceed to the fully-connected layers.
To reduce this dimensionality and computation cost, MaxPooling layers were introduced after every convolutional layer, each parametrized with a kernel size and stride of 2. \newline
With regard to the size of the feature vector after the last pooling layer, it was decided to incorporate two fully-connected layers for the final dimensionality reduction, with the first one halving the signal length, before the concluding layer reduced it to a single parameter.\newline
As the chosen pre-processing measure does not scale the pressure data to commonly used ranges as [-1,1] or [0,1], the ReLU (Rectified Linear Unit) function was chosen as activation function for every layer but the final one, as it enables effective learning even without a normalized input signal \cite{KRIZHEVSKY:12}.
To obtain the final knock probability value predicted by the model, the resulting, single-dimensional parameter was scaled using a Sigmoid activation function. A graphical representation of the fundamental model architecture is illustrated in Fig. \ref{fig:cnn_arch}.\newline
The choice of a convolutional neural network allows for effective processing of the signal with a lower number of parameters in opposition to completely fully-connected networks, thereby also being more robust against the vanishing gradient problem commonly encountered when dealing with Recurrent Neural Networks (RNNs).
Furthermore, the network allows for a high degree of parallelism, greatly increasing the computation speed in comparison to both of the aforementioned neural network architectures \cite{FAWAZ:19}.
Table \ref{tab:total_parameters} gives an overview of the four models designed for this study.

\begin{table}[!ht]
    \centering
    \small
    \caption{Number of learnable parameters and total size per model}
    \resizebox{0.45\textwidth}{!}{%
    \begin{tabular}{l c c c c}
    \toprule
         & \textbf{Model a} & \textbf{Model b} & \textbf{Model c} & \textbf{Model d} \\  
         \textbf{Target frequency} & 3 kHz & 3.9 kHz & 4.9 kHz & 8.1-8.7 Hz \\ \midrule 
         \textbf{Layer}& \multicolumn{4}{c}{\textbf{Parameters (Channels)}} \\ 
         Conv1 & 30 (4) & 23 (4) & 18 (4) & 11 (4) \\ 
         MaxPool & 0 & 0 & 0 & 0 \\ 
         Conv2 & 30 (8) & 23 (8) & 18 (8) & 11 (8) \\ 
         MaxPool & 0 & 0 & 0 & 0 \\
         Conv3 & 61 (16) & 47 (16) & 37 (16) & 23 (16) \\
         MaxPool & 0 & 0 & 0 & 0 \\
         FC1 & 226,128 & 307,720 & 446,040 & 609,960 \\
         FC2 & 337 & 393 & 473 & 553 \\
         Sigmoid & 0 & 0 & 0 & 0 \\
         \textbf{TOTAL} & \textbf{227,801} & \textbf{309,141} & \textbf{447,321} & \textbf{611,013} \\ 
         \midrule
         \textbf{Model size [MB]} & 45.0 & 61.3 & 88.8 & 121.4 \\ 
         \bottomrule
    \end{tabular}%
    }
    \label{tab:total_parameters}
\end{table}

\subsection{Training \& Evaluation}

All CNN models were trained in supervised manner, using the aforementioned set of "relative" labels as the ground truth for training. Hence, the learning process is based on a multi-category knock label.
Relative labels were scaled to a range between 0 and 1 to ensure compatibility with the chosen Binary-Cross-Entropy loss function for the neural network.
Separated from training, probabilities output from the model were also compared to the "binary" label set after first applying an equivalent conversion.\newline
Hence, models were evaluated with two main metrics: (i) a binary accuracy, indicating how well the CNN could separate knocking cycles from non-knocking cycles and (ii) a multi-categorical accuracy showing the models' performance when classifying the severity of knocking cycles. While the former method can be summarized in one accuracy value per analysis, the latter is more effectively depicted using confusion matrices. 
Furthermore, since the CNN outputs probabilites on a continuous scale from 0 to 1, a conversion process was applied to fit the six classes (0 votes to 5 votes). 
The conversion key is illustrated in Table \ref{tab:rel_convert}.

\begin{table}[!ht]
    \centering
    \caption{Conversion from model output probabilities to relative label}
    \label{tab:rel_convert}
    \resizebox{0.45\textwidth}{!}{%
    \begin{tabular}{l|c|c|c|c|c|c}
    \toprule
         \textbf{Converted label} & \textbf{0} & \textbf{1} & \textbf{2} & \textbf{3} & \textbf{4} & \textbf{5} \\
         \midrule
         \textbf{Probability range} & $<$ 0.1 & 0.1 - 0.3 & 0.3 - 0.5 & 0.5 - 0.7 & 0.7 - 0.9 & $>$ 0.9 \\
         \bottomrule 
    \end{tabular}%
    }
\end{table}

For training, the following hyperparameters collected in Table \ref{tab:hyperparams} were chosen by executing a grid search using the 70/70/70 split used in the main analysis.

\begin{table}[!ht]
    \centering
    \footnotesize
    \caption{Model hyperparameters}
    \begin{tabular}{lr}
    \toprule
        \multicolumn{2}{c}{\textbf{Hyperparameters}} \\
        \midrule
         Initial learning rate & 1e-3 \\
         Batch Size & 64 \\
         Epochs (max.) & 200 \\
         Regularization penalty & 1e-4 \\ 
         \midrule
         MaxPool (size, stride) & (2, 2) \\
         Regularization & L2 \\
         Loss function & Binary Cross-Entropy \\
         Optimizer & Adam \\
         \bottomrule
    \end{tabular}
    \label{tab:hyperparams}
\end{table}

\noindent The training process for each model was tuned for early stopping once a plateau in classification accuracy was achieved or diverging trends in train and test accuracies, i.e. overfitting was detected.

\section{Results \& Discussion}
\label{sec:result}

As mentioned above, all models described in this section used a 70/70/70 train-test-split - unless explicitly noted otherwise.

\subsection{Reference methods}

To measure the models' performance, three methods were chosen as reference, the first of which being the MAPO criterion. 
MAPO is a widely used detection method deployed in the test benches that had been used to obtain the training data for this work. 
It is an example of a TVE technique, which distinguishes between knocking and non-knocking cycles by comparing the maximum amplitude of pressure oscillations (\eqref{eq:mapo-calc}) of a band pass filtered signal to a pre-defined threshold value $t$ (\eqref{eq:mapo-comp}).

\begin{equation}
    label_i = 
    \begin{cases}
        (1)\: \text{knocking}, & MAPO_i > t \\
        (0)\: \text{non-knocking}, & MAPO_i < t
    \end{cases}
    \label{eq:mapo-comp}
\end{equation}

\noindent with

\begin{equation}
    MAPO_i = max\left| \: p_{filt, i}(\theta) \: \right|
    \label{eq:mapo-calc}
\end{equation}

As evident from the definition, the procedure is highly dependent on $t$.
For reliable detection, the threshold placement demands thorough calibration to the evaluated engine and present operating conditions. 
In this study, however, instead of placing a rigid MAPO threshold value for all cycles, an optimum value for each train-test-split has been calculated via logistic regression. 
This enables a comparison not only to an optimized MAPO criterion, but also to this tuned criterion's generalization ability when applying optimum values from one set to another.\newline
It must be noted that this MAPO tuning is only possible as a post-operation process. 
The accuracy displayed in this paper is therefore not realistic for actual test-bed experiments.
Furthermore, MAPO threshold values are typically submitted by engine manufacturers and are in part set to avoid possibly damaging knocking conditions altogether. \newline
The other two reference methods were reproduced from \cite{PANZANI:19} and represent criteria based on a PCA of the in-cylinder pressure signal - see Section \ref{sec:related}. 
The first approach uses a defined number of the calculated principal components to reconstruct the pressure signal, followed by the extraction of two distinct features: (i) the RMSE error between the original signal and the reconstruction, and (ii) the maximum amplitude (MAPO) of the residual pressure curve, which is obtained by subtracting the smoother reconstruction signal from the original. 
The thus obtained values are then classified using logistic regression. 
The second approach, labeled as the purely data-driven procedure, uses the inner products between the pressure trace under consideration and the principal components as input for the logistic regression classification. 
This study will use the abbreviations "PCA Eigen" for the former method and "PCA DD" for the latter.\newline
The input data used for the reference methods' evaluation is identical to the 60° CA window used for training and testing the CNN model.

\subsection{Individual model performance analysis}

To ensure the model performance's validity, a tenfold cross-validation was conducted, the results of which are captured in Table \ref{tab:main_analysis} in terms of the binary accuracy metric. 
The best performing models have been highlighted for each of the evaluation criteria at the bottom of the table. 
All values represent a model's accuracy and derived statistic quantities, respectively.

\begin{table}[!ht]
    \centering
    \caption{Individual model performance results in terms of binary accuracy - best performing models are highlighted}
    \resizebox{0.48\textwidth}{!}{%
    \begin{tabular}{r|cc|cc|cc|cc}
        \toprule
         & \multicolumn{2}{c|}{\textbf{Model a}} & \multicolumn{2}{c|}{\textbf{Model b}} & \multicolumn{2}{c|}{\textbf{Model c}} & \multicolumn{2}{c}{\textbf{Model d}} \\
         \textbf{\#} & Train & Test & Train  & Test & Train & Test & Train & Test \\
         \midrule
        1 & 0.9390 & 0.9293 & 0.9638 & 0.9432 & 0.9712 & 0.9374 & 0.9474 & 0.9421  \\
        2 & 0.9479 & 0.9386 & 0.9688 & 0.9235 & 0.9588 & 0.9339 & 0.9623 & 0.9247  \\
        3 & 0.9648 & 0.9444 & 0.8879 & 0.8864 & 0.9584 & 0.9409 & 0.9504 & 0.9397  \\
        4 & 0.9276 & 0.9247 & 0.9593 & 0.9409 & 0.9474 & 0.9455 & 0.9648 & 0.9351  \\
        5 & 0.9449 & 0.9409 & 0.9449 & 0.9409 & 0.9588 & 0.9420 & 0.9276 & 0.9247  \\
        6 & 0.9707 & 0.9328 & 0.9688 & 0.9316 & 0.9524 & 0.9351 & 0.9638 & 0.9397  \\
        7 & 0.9043 & 0.9027 & 0.9529 & 0.9328 & 0.9727 & 0.9409 & 0.9425 & 0.9351  \\
        8 & 0.9474 & 0.9363 & 0.9405 & 0.9363 & 0.9722 & 0.9421 & 0.9455 & 0.9397  \\
        9 & 0.9419 & 0.9305 & 0.9762 & 0.9189 & 0.9167 & 0.9131 & 0.9554 & 0.9397  \\
        10 & 0.9668 & 0.9351 & 0.9717 & 0.9374 & 0.9479 & 0.9397 & 0.9618 & 0.9409  \\
         \midrule
        Mean   & 0.9456 & 0.9315 & 0.9535 & 0.9292 & \textbf{0.9557} & \textbf{0.9371} & 0.9522 & \textbf{0.9362} \\
        Median & 0.9462 & 0.9339 & \textbf{0.9616} & 0.9345 & 0.9586 & \textbf{0.9403} & 0.9529 & \textbf{0.9397} \\
        Max    & 0.9707 & 0.9444 & \textbf{0.9762} & 0.9432 & 0.9727 & \textbf{0.9455} & 0.9648 & 0.9421 \\
        Min    & 0.9043 & 0.9027 & 0.8879 & 0.8864 & 0.9167 & 0.9131 & \textbf{0.9276} & \textbf{0.9247} \\
        std-dev & 0.0188 & 0.0111 & 0.0245 & 0.0160 & 0.0158 & 0.0086 & \textbf{0.0113} & \textbf{0.0061} \\
        \bottomrule
    \end{tabular}%
    }
    \label{tab:main_analysis}
\end{table}

\begin{figure}[!ht]
    \centering
    \includegraphics[width=0.36\textwidth, trim={3.5cm 0.8cm 2cm 0.8cm
    0},clip]{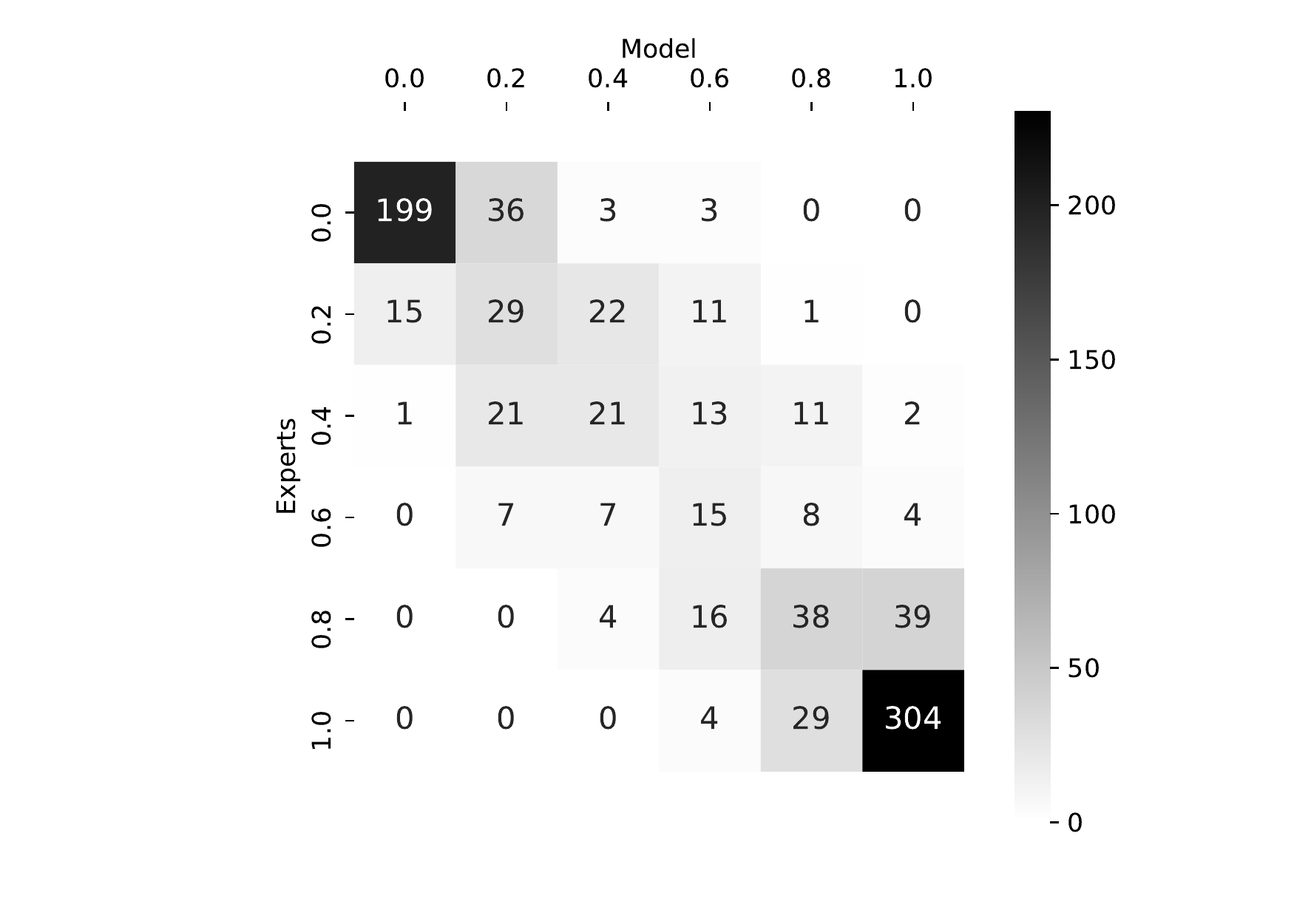}
    \caption{Confusion Matrix for one trained instance of Model $d$. The main diagonal shows the cycles which have been perfectly classified, while the secondary diagonals show the cycles which the model placed one category above/below the expert rating. Ratings from the relative label set are scaled to a [0;1] range.}
    \label{fig:confmat}
\end{figure}

\noindent First and foremost, results on the test sets show that Model $d$ ($f_t = 7.8-8.7$ kHz) performance on the test set ranks at the top in all but one category.\newline
An observation of the CNN-based models' results here allows for an evaluation between the different target frequencies. 
Model $a$ ($f_t = 3$ kHz) and Model $b$ ($f_t = 3.8-4.0$ kHz) show demonstrative performance in fitting the training data, but performance on the test set cannot match the other two models.
Both Models $c$ ($f_t = 4.8-5.2$ kHz) and $d$ show convincing results in all categories. 
However, the latter's minimum accuracy among all folds of the cross-validation is more than 1\% higher as compared to the former.
Furthermore, Model $d$ also offers the most consistent test results, as the standard deviation is the lowest among all approaches in this work. 
Furthermore, each CNN model achieved a classification within 1 ms, with the fastest architecture reaching an average time of $\sim0.2$ ms per cycle. \newline
Considering the accuracy on the set of relative labels, as mentioned above, confusion matrices were employed for evaluation. Fig. \ref{fig:confmat} shows one of these matrices. 

Since not all of the matrices can be shown in this paper, Table \ref{tab:conf_diag} gives an overview of the results achieved in the multi-category classification. 
The first row for each instance gives the accuracy across the main diagonal of the confusion matrix, concretely the rate of perfectly classified cycles. 
As can be observed, the most common errors are misclassifications between label pairs 0.0-0.2 or 0.2-0.4, respectively (upper left), and 0.8-1.0 (lower right).
However, by the label definitions in Section \ref{sec:dataset}, this solely reflects imprecision when judging the severity of the knocking event.
In the former two cases, this is irrelevant for the engine control use-case, as the model still correctly classifies each of the concerned cycles among the "non-damaging" categories.
In the latter case, a confusion of the labels "knock" and "severe knock" is also judged an insignificant error in regard of the mentioned use-case.
Accordingly, the second row in Table \ref{tab:conf_diag} illustrates the accuracy if the secondary diagonal is added to the number of correctly classified cycles.
The third row considers missing by one position as correct, unless this would change the converted binary label, e.g. a scaled relative label of 0.4 (binary "non-knocking") classified as 0.6 (binary "knocking").
All values given represent the mean values of the tenfold cross-validation results on the test set. 

\begin{figure}[!ht]
    \centering
    \begin{tikzpicture}
        \begin{axis}[
                width=7.8cm,
                height=3.5cm,
                at={(0,0)}, 
                ymin=0, 
                ymax=0.00009, 
                xmin=0, 
                xmax=40, 
                grid=both, 
                major grid style={thin,color=black!10},
                xtick={0,5,...,40}, 
                yticklabels={,,},
                scaled y ticks = false,
                ]
        \addplot[color=black,smooth] table [x=a, y=c1, col         sep=semicolon]{fig/FIG6_src_A.txt}; 
        \node[draw] at (5.25cm, 1.5cm) {Model a};
        \end{axis}
    \end{tikzpicture}
    \begin{tikzpicture}
        \begin{axis}[
                width=7.8cm, height=3.5cm, 
                at={(0,0)}, ymin=0, ymax=0.000015, xmin=0, xmax=40, grid=both, major grid style={thin,color=black!10}, xlabel={Frequency [kHz]}, xtick={0,5,...,40}, yticklabels={,,}, scaled y ticks = false,
                ]
        \addplot[color=black,smooth] table [x=a, y=c4, col     sep=semicolon]{fig/FIG6_src_B.txt};
        \node[draw] at (5.25cm, 1.5cm) {Model d};
        \end{axis}
    \end{tikzpicture}
    \caption{FFT analysis of CNN coefficients}
    \label{fig:fft}
\end{figure}
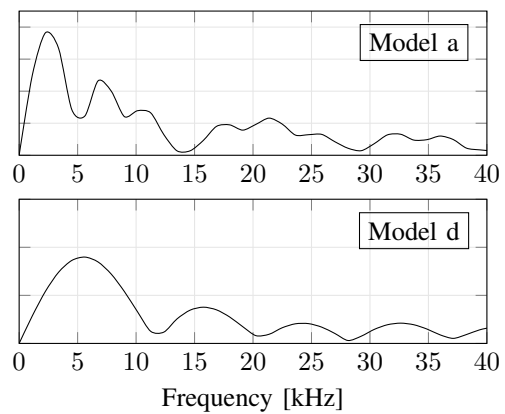

\begin{table}[!ht]
    \centering    
    \caption{Translation of the confusion matrix results for multi-category classification}
    \resizebox{0.48\textwidth}{!}{%
    \begin{tabular}{l|cccc}
    \toprule
         Base of evaluation & Model $a$ & Model $b$ & Model $c$ & Model $d$ \\
         \midrule
         Main diagonal & 0.6845 & 0.6905 & 0.7016 & \textbf{0.7034} \\
         Main + secondary & 0.9317 & 0.9257 & 0.9328 & \textbf{0.9384} \\
         Main + secondary (mod.) & 0.9027 & 0.8969 & 0.9049 & 
         \textbf{0.9099} \\
         \bottomrule
    \end{tabular}%
    }
    \label{tab:conf_diag}
\end{table}
\noindent Regarding the evaluation of the multi-category classification, it can again be noted that while all individual models have a similar performance, Model $c$ and $d$ again show the best results. 
The high accuracy levels of above 90\% in the third row of the table, combined with an analysis of the confusion matrices (as in Fig. \ref{fig:confmat}) show remarkable performance.
Especially conditions of severe knock and consensually rated normal combustion are estimated at high precision. 
Furthermore, grave misclassifications are almost non-existent, as the models do not give high ratings to cycles with low expert labels and vice-versa.\newline
To evaluate the models' underlying hypothesis of frequency-based architecture, each model instance's first layer coefficients were extracted from the model. 
Since it was expected that this layer would adapt to and amplify a certain frequency, the coefficients were processed using a Fast-Fourier-Transform (FFT), with the assumption that there would be one dominant peak in the resulting spectrum. 
Fig. \ref{fig:fft} shows the FFT results for the best-performing instances of Model $a$ and $d$. 

\noindent The FFT results confirm the expectation that distinct peaks would be recognizable when analysing the spectrum obtained from the model coefficients. 
It is therefore concluded that a sinusoidal pattern with a dominant component is learned by the CNN's first layer. 
An investigation of Fig. \ref{fig:fft} shows that both models' learned spectrum exhibits increased amplitudes in the same regions, with Model $d$ essentially representing a lower resolution version of Model $a$. 
This can directly be traced back to the respective first layers' number of parameters, specifically 30 for Model $a$ and 11 for Model $d$. 
Furthermore, it can be assumed as the reason for each models' similar performance: while the neural network is able to extract the relevant frequency-related features in each instance, the kernel size determines the resulting spectrum's resolution and with it, the performance. 
With the distinct engines' resonance frequencies scattered over a certain range, it can be concluded that the wider peak exhibited in the Model $d$ FFT is more beneficial for generalization processes, due to its amplification of a likewise broader range of frequencies. 
In addition, a higher number of parameters in individual filters increases the potential to overfit.
The spectrum observed for increased kernel size might therefore be tuned to the specific frequencies observed in cycles from the training data, in turn diminishing the performance on other cycles in the test set.
Thus, the superior level of performance of smaller kernel models can be explained.

\subsection{Model performance against reference methods}

With Model $d$ showing the highest accuracy paired with the most stable results among the proposed architectures for both metrics, its performance was measured against the above-mentioned reference methods. 
Again, a tenfold cross-validation was conducted for all procedures. 
All reference methods were optimized. 
Hence, the chosen MAPO threshold was the best possible for each train-test-split and both of the illustrated PCA-based methods represent the respective best performing model.
In either case, this proved to be the instance built on the extraction of eight principal components.

\begin{table}[!ht]
    \centering
    \caption{Comparison to reference methods}
    \resizebox{0.48\textwidth}{!}{%
    \begin{tabular}{c|cc|cc|cc|cc}
    \toprule
         & \multicolumn{2}{c|}{\textbf{Model d}} & \multicolumn{2}{c|}{\textbf{MAPO TVE}} & \multicolumn{2}{c|}{\textbf{PCA DD}} & \multicolumn{2}{c}{\textbf{PCA Eigen}} \\
         \textbf{\#} & Train & Test & Train  & Test & Train & Test & Train & Test\\
         \midrule
         1 & 0.9474 & 0.9421 & 0.8648 & 0.8621 & 0.8586 & 0.8462 & 0.8476 & 0.8613 \\
         2 & 0.9623 & 0.9247 & 0.8699 & 0.8580 & 0.8496 & 0.8462 & 0.8452 & 0.8647 \\
         3 & 0.9504 & 0.9397 & 0.8777 & 0.8667 & 0.8486 & 0.8273 & 0.8476 & 0.8555 \\
         4 & 0.9648 & 0.9351 & 0.8623 & 0.8751 & 0.8471 & 0.8555 & 0.8506 & 0.8474 \\
         5 & 0.9276 & 0.9247 & 0.8646 & 0.8654 & 0.8347 & 0.8566 & 0.8521 & 0.8462 \\
         6 & 0.9638 & 0.9397 & 0.8636 & 0.8701 & 0.8516 & 0.8370 & 0.8556 & 0.8335 \\
         7 & 0.9425 & 0.9351 & 0.8656 & 0.8698 & 0.8452 & 0.8601 & 0.8496 & 0.8451 \\
         8 & 0.9455 & 0.9397 & 0.8701 & 0.8688 & 0.8546 & 0.8301 & 0.8551 & 0.8347 \\
         9 & 0.9554 & 0.9397 & 0.8543 & 0.8540 & 0.8442 & 0.8555 & 0.8536 & 0.8393 \\
         10 & 0.9618 & 0.9409 & 0.8698 & 0.8489 & 0.8536 & 0.8301 & 0.8551 & 0.8405 \\
         \midrule
         Mean & \textbf{0.9522} & \textbf{0.9362} & 0.8663 & 0.8639 & 0.8488 & 0.8442 & 0.8512 & 0.8468 \\
         Median &  \textbf{0.9529} & \textbf{0.9397} & 0.8652 & 0.8661 & 0.8491 & 0.8462 & 0.8514 & 0.8457 \\
         Max & \textbf{0.9648} & \textbf{0.9421} & 0.8777 & 0.8751 & 0.8586 & 0.8601 & 0.8556 & 0.8647 \\
         Min & \textbf{0.9276} & \textbf{0.9247} & 0.8543 & 0.8489 & 0.8347 & 0.8242 & 0.8452 & 0.8335 \\
         std-dev & 0.0113 & \textbf{0.0061} & \textbf{0.0062} & 0.0081 & 0.0066 & 0.0130 & 0.0037 & 0.0107 \\
         \bottomrule 
    \end{tabular}%
    }
    \label{tab:comp_ref}
\end{table}

Table \ref{tab:comp_ref} shows that applied to the cycles in this study, Model $d$ outperforms all reference methods by a significant margin of at least 7.2\% when regarding the cross-validation's mean test accuracy. 
Furthermore, the proposed CNN architecture provides a better fit to training data across all different train-test-splits. 
Generally, the presented model outperforms the reference methods in every category with the exception of the standard deviation on training data fit. \newline
For further evaluation, Model $d$ was selected to analyze performance when generalizing on data from unseen engines and after training on smaller datasets.

\subsection{Generalization to unseen engines}

For the purpose of observing the transferability of extracted features from one engine to another, the best-performing model from the main analysis was retrained using training set compositions entirely omitting at least one of the subsets. 
It was expected that overfitting would occur significantly earlier when training with these reduced datasets, however this was prevented by again implementing early stopping once the trend was observed.
As a reference, the MAPO criterion was again optimized via logistic regression. Additionally, both PCA methods were applied to the newly designed train and test sets.
An overview of all investigated combinations, as well as results for binary accuracies are captured in Table \ref{tab:gen_app}. 

\begin{table*}[!ht]
    \centering
    \caption{Set composition and results for application of learned features to unseen engines - best performing models are highlighted}
    \resizebox{0.72\textwidth}{!}{%
    \begin{tabular}{c|cc|cc|cc|cc|cc}
    \toprule
        & \multicolumn{2}{c|}{\textbf{Subsets in}} & \multicolumn{2}{c|}{\textbf{Model $d$}} & \multicolumn{2}{c|}{\textbf{MAPO (optim.)}} &
        \multicolumn{2}{c|}{\textbf{PCA DD}} & \multicolumn{2}{c}{\textbf{PCA Eigen}}\\
        \textbf{\#} & Train set & Test set & Train acc. & Test acc. & Train acc. & Test acc. & Train acc. & Test acc. & Train acc. & Test acc. \\
         \midrule
        1 & A & BC & \textbf{0.9301} & \textbf{0.8607} & 0.9214 & 0.7644 & 0.8905 & 0.7735 & 0.8095 & 0.7569 \\
        2 & B & AC & 0.9187 & \textbf{0.8059} & \textbf{0.9353} & 0.7936 & 0.8673 & 0.6638 & 0.8980 & 0.7261 \\
        3 & C & AB & 0.8786 & \textbf{0.7295} & \textbf{0.9333} & 0.5692 & 0.8667 & 0.6650 & 0.9259 & 0.4996 \\
        4 & AB & C & 0.8462 & \textbf{0.7503} & \textbf{0.9316} & 0.6481 & 0.8389 & 0.7074 & 0.8761 & 0.6185 \\
        5 & AC & B & \textbf{0.9359} & \textbf{0.9027} & 0.8159 & 0.6380 & 0.8717 & 0.7813 & 0.7906 & 0.5307 \\
        6  & BC & A & \textbf{0.9546} & \textbf{0.9085} & 0.8441 & 0.8738 & 0.8554 & 0.6357 & 0.8074 & 0.8214 \\
        \bottomrule
    \end{tabular}%
    }
    \label{tab:gen_app}
\end{table*}

\noindent First observations of the results show that the CNN again outperforms the reference methods in all test set related accuracy measures.
While the MAPO criterion shows better fits on the training data in three of the scenarios, this can be directly traced back to the early stopping of the CNN.
It can therefore be assumed that the CNN would have achieved a better fit, albeit deteriorating the test set performance in the process.
The considerable margins between MAPO training and test fit, especially in scenarios 3 and 4 highlight the method's need to be calibrated precisely to individual engines. 
Both PCA methods do not reach comparable levels of training data fit, with the exception of the PCA Eigen approach reaching a train accuracy of 92,6\% in scenario 3. \newline
From a more detailed analysis of the results, it can be inferred that features from subsets A and B do not translate well to the different engine type subset C. 
This is emphasized by the fact that in scenarios 3 and 4 none of the methods achieve scores of over 72\% or 75\%, respectively.\newline 
By contrast, the incorporation of different engine types in the training set shows almost no deterioration in classifying accuracy. 
As can be obtained from scenarios 5 and 6, the CNN fits the distinct engines adequately, while being able to transfer the extracted features considerably well to another engine - in turn outperforming references by at least 12\% (scenario 5).\newline
With the exception of the well-performing scenarios 5 and 6, all cases show a sizeable margin between train and test loss, indicating overfitting and/or a covariance shift due to the different engine characteristics. 
To investigate how much input data is needed from distinct engines to uphold performance and minimize overfitting, an analysis on training with small datasets was conducted.

\subsection{Training on small data fractions}

Training sets in this analysis included knocking and non-knocking cycles from each subset, however, their number was continuously reduced.
Furthermore, the number of cycles from subset C was decreased at a greater rate, to investigate the effect on feature transferability observed in the previous section.
The results of this process are captured in Table \ref{tab:smallmod} and again represent the mean binary accuracy of a tenfold cross-validation. 
Test accuracies for each case are given separately for the test cycles from each respective subset.
Again, no hyperparameters were modified for this analysis and early stopping was activated to prevent severe overfitting.
\noindent This analysis' results show that the CNN approach is able to classify with considerable accuracy even when trained on remarkably small fractions of data, as long as data from each engine is available for training. 
This confirms the conclusions made in the previous subsection, displaying that the CNN is able to efficiently extract shared features from distinct engines and operating conditions.\newline
To provide a possible methodology for enhancing the application of these findings, the last split was trained using 20\% of only non-knocking cycles from engine C, respectively. 
As a result, the accuracy on subset C test cycles increased by 1.5\% when compared to the standard 50/50/20 split, while accuracy on the other subsets is hardly influenced.
This suggests that the CNN's ability to transfer features can be boosted considerably by adding non-knocking cycles of new engines to the training set.
Acquiring these cycles is easily achieved by operating the engine outside of the knock limit.
Therefore, the extensive calibration process when handling a new engine with common detection approaches can be replaced with minor training data extensions.\newline
The exhibited performance is also notable regarding another aspect of the training set's composition. 
Due to the drastically reduced number of cycles and the randomized split, there is a high probability that at least one operating point's data is entirely omitted for each engine in several of the cross-validation splits. 
As a consequence, it can be suggested that the model is able to generalize well to unseen operating conditions, though definite confirmation would require tests on different data.

\begin{table}[!ht]
    \centering
    \caption{Model d performance on small datasets\newline
    *only non-knocking cycles}
    \resizebox{0.41\textwidth}{!}{%
    \begin{tabular}{c|ccc|c|ccc}
    \toprule
         & \multicolumn{3}{c|}{\textbf{Training cycles per subset}} & \textbf{Train acc.} & \multicolumn{3}{c}{\textbf{Test acc. per subset}} \\
         \textbf{Split} & A & B & C & & A & B & C\\
         \midrule
         70/70/0 & 588 & 1050 & 0 & 0.9444 & 0.9381 & 0.9438 & 0.7503 \\
         \midrule
         50/50/20 & 420 & 750 & 108 & 0.9357 & 0.9246 & 0.9264 & 0.8748 \\
         35/35/10 & 294 & 525 & 54 & 0.9386 & 0.9269 & 0.9337 & 0.8786 \\
         25/25/8 & 210 & 375 & 43 & 0.9379 & 0.9254 & 0.9298 & 0.8497 \\
         15/15/5 & 126 & 225 & 27 & 0.9261 & 0.9178 & 0.9102 & 0.8329 \\
         \midrule
         50/50/20* & 420 & 750 & 67 & 0.9325 & 0.9211 & 0.9275 & 0.8902 \\
         \bottomrule
    \end{tabular}%
    }
    \label{tab:smallmod}
\end{table}

\section{Conclusion}
\label{sec:conclusion}

This paper introduces a 1D convolutional neural network approach for knock detection in a spark ignition combustion engine with the network's layers designed to capture frequency-dependent features.
Being trained on data from numerous operating points of three distinct engines, the model showed considerable classification accuracy.
A binary distinction between "knocking" and "non-knocking" cycles yielded consistent results of over 92\% accuracy, while for a multi-class problem, the model classified 78\% of cycles perfectly and over 90\% at most one position from ground truth.
Thus, the proposed CNN models outperform the widely used MAPO test bench knock criterion, as well as two PCA-based criteria by a significant margin in all experiments conducted for this study. 
Cycles were cut to a 60° CA window starting at TDC position, without any further scaling or pre-processing.\newline
Extensive analyses of the acquired test results showed that the frequency-based design proved to work by extracting sinusoidal patterns from the input signals. 
Analysing these patterns' frequency spectra provided an insight into the neural network's learning process, also allowing conclusions on the reasons for each model's respective performance. 
Therefore, the approach can be described as a successful implementation of theory-guided data science using physics-inspired relations in the CNN architecture design.\newline 
It was further found that while models trained on only one or similar types of engine showed difficulties in generalizing to unseen engines, a minimal amount of non-knocking cycles from another engine in the training data was enough to boost feature transferability.
Additionally, it was illustrated via small dataset training that highly efficient generalization to different operating points is also possible.\newline
This enhanced generalization ability presents a considerable advantage when compared to other  knock detection approaches. 
With a classification speed of under 1 ms per cycle and a manageable model size, the method is furthermore capable of real-time classification and incorporation into test bench operation.\newline
The authors' future work will concentrate on expanding the proposed knock detection approach to construct a condition forecasting model, effectively pursuing the prediction of knock occurrences in upcoming combustion cycles.

\appendices

\section*{Acknowledgment}

The authors acknowledge the financial support of the Austrian COMET - Competence Centers for Excellent Technologies - Programme of the Austrian Federal Ministry for Climate Action, Environment, Energy, Mobility, Innovation and Technology, the Austrian Federal Ministry for Digital and Economic Affairs, and the States of Styria, Upper Austria, Tyrol, and Vienna for the COMET Centers Know-Center and LEC EvoLET, respectively. The COMET Programme is managed by the Austrian Research Promotion Agency (FFG).

\ifCLASSOPTIONcaptionsoff
  \newpage
\fi

\bibliographystyle{IEEEtran}
\bibliography{Transactions-Bibliography/bibliography}

\begin{IEEEbiography}[{\includegraphics[width=1in,height=1.25in,clip,keepaspectratio]{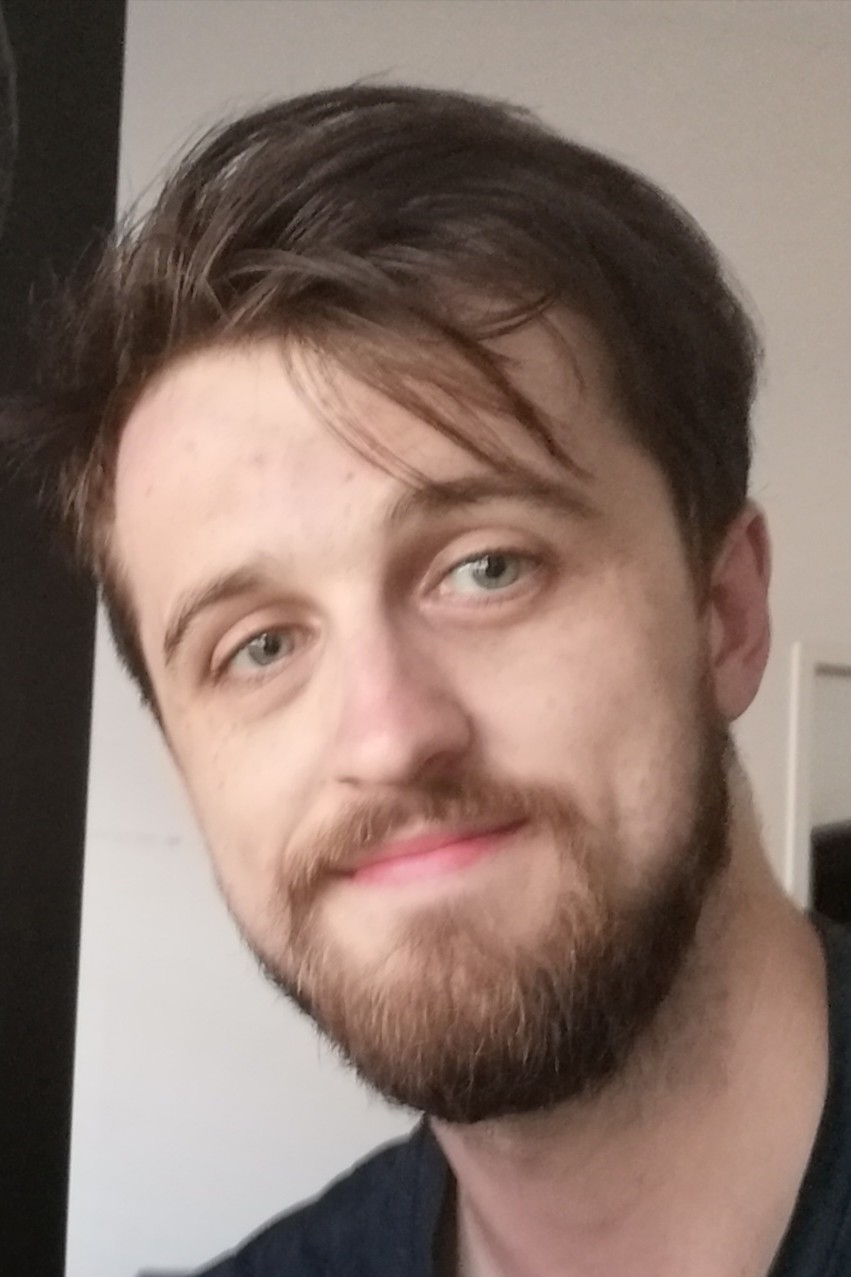}}]{Andreas B. Ofner}
received his Master of Science degree in mechanical engineering from Carinthia University of Applied Science in 2018. \newline
Subsequently, he worked at AVL, an international developer of powertrain systems based in Graz, Austria, for 2 years as a simulation engineer in the NVH department. He is currently a PhD candidate at Know-Center GmbH in Graz, Austria, studying combustion phenomena using methodologies from the data science \& artificial intelligence domain.
\end{IEEEbiography}

\begin{IEEEbiography}[{\includegraphics[width=1in,height=1.25in,clip,keepaspectratio]{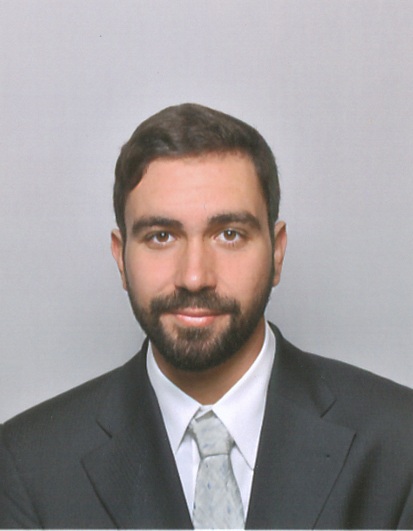}}]{Achilles Kefalas}
received his Master of Science degree in mechanical engineering from Graz University of Technology in 2013. Subsequently, he worked at Andritz AG, an international developer of hydraulic turbomachinery based in Graz, Austria for five years. He is currently a PhD candidate at the Institute of Internal Combustion Engines and Thermodynamics of Graz University of Technology, studying combustion phenomena using methodologies from data science, artificial intelligence as well as thermodynamics domains.\end{IEEEbiography}

\begin{IEEEbiography}[{\includegraphics[width=1in,height=1.25in,clip,keepaspectratio]{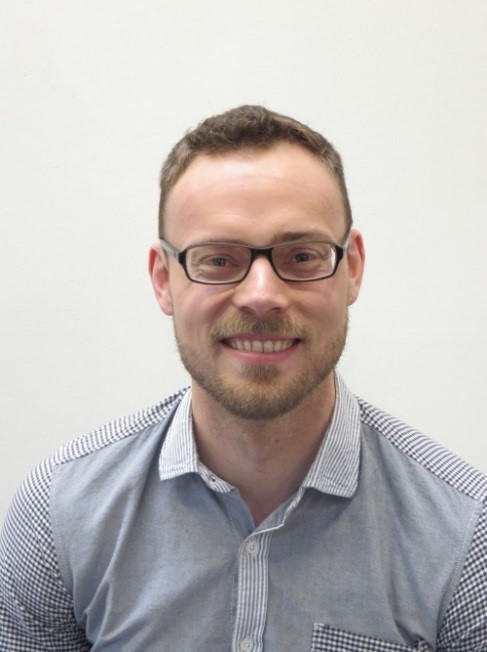}}]{Stefan Posch}
received the B.Sc., M.Sc. and Ph.D. in mechanical engineering at Graz University of Technology, Austria, in 2011, 2013 and 2017, respectively. He then worked as a senior engineer at Midea Austria GmbH where he was responsible for simulation tasks in the field of hermetic compressors. Since 2019 he has been working at the Large Engines Competence Center GmbH in Graz, Austria, as a senior scientist and team leader for system simulation and AI integration. His main research interests include the combination of numerical simulation and data-driven approaches.  \end{IEEEbiography}

\begin{IEEEbiography}[{\includegraphics[width=1in,height=1.25in,clip,keepaspectratio]{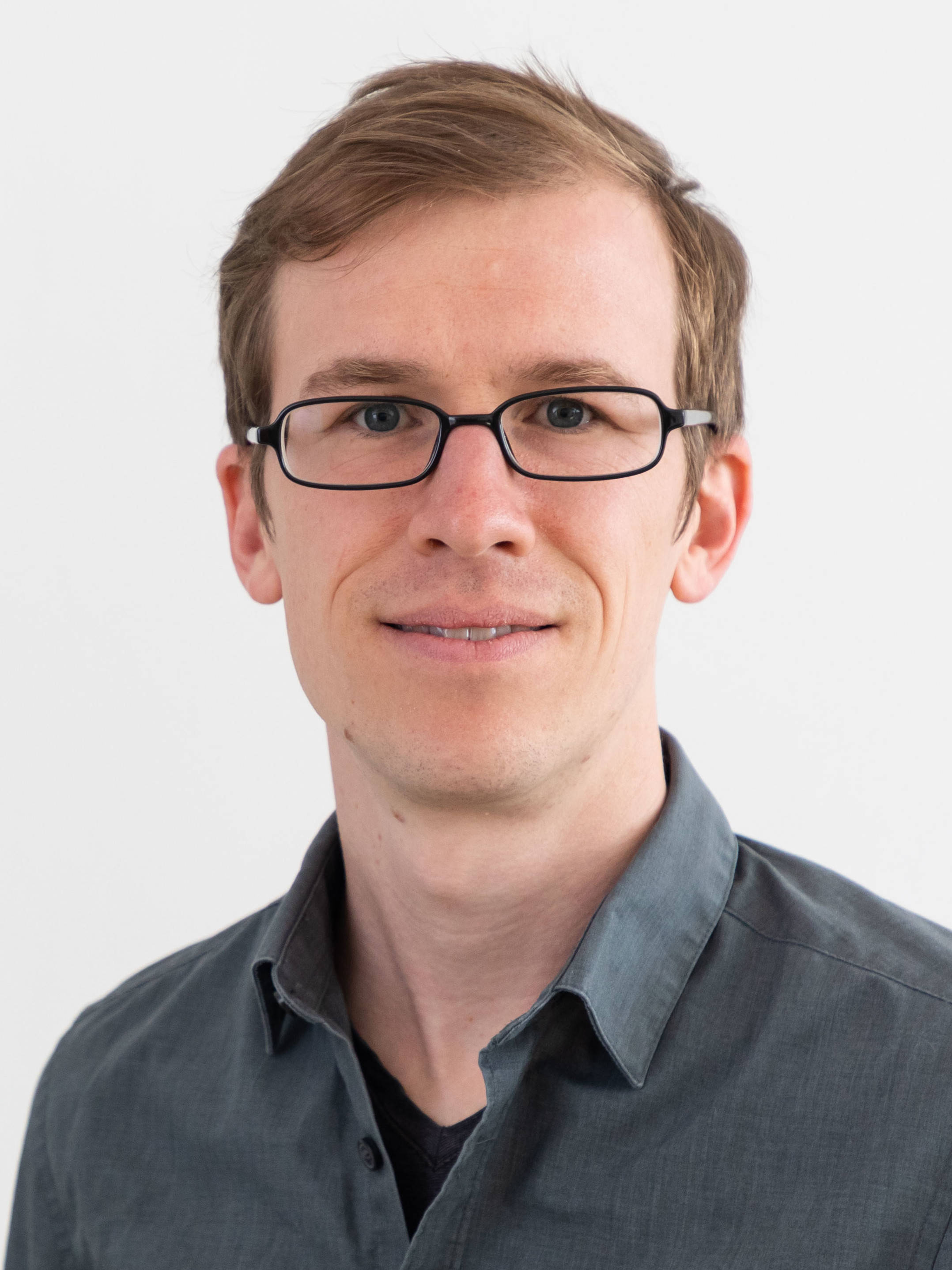}}]{Bernhard C. Geiger}
received the Dipl.-Ing. degree in electrical engineering (with distinction) and the Dr. techn. degree in electrical and information engineering (with distinction) from Graz University of Technology, Austria, in 2009 and 2014, respectively.\newline
In 2010, he joined the Signal Processing and Speech Communication Laboratory, Graz University of Technology, as a Research and Teaching Associate. He was a Senior Scientist and Erwin Schr\"odinger Fellow at the Institute for Communications Engineering, Technical University of Munich, Germany from 2014 to 2017 and a postdoctoral researcher at the Signal Processing and Speech Communication Laboratory, Graz University of Technology, Austria from 2017 to 2018. He is currently a Senior Researcher at Know-Center GmbH, Graz, Austria, where he leads the Machine Learning Group within the Knowledge Discovery Area. His research interests cover information theory for machine learning, theory-assisted machine learning, and information-theoretic model reduction for Markov chains and hidden Markov models
\end{IEEEbiography}

\vfill

\end{document}